\documentclass[10pt,twocolumn,letterpaper]{article}

\usepackage{iccv}
\usepackage{times}
\usepackage{epsfig}
\usepackage{graphicx}
\usepackage{amsmath}
\usepackage{amssymb}
\usepackage{algorithm}
\usepackage{multirow}
\usepackage[noend]{algpseudocode}
\usepackage{enumitem}
\usepackage{caption} 
\usepackage[pagebackref=true,breaklinks=true,letterpaper=true,colorlinks,bookmarks=false]{hyperref}

\iccvfinalcopy % *** Uncomment this line for the final submission

\def\iccvPaperID{3795} % *** Enter the ICCV Paper ID here

\ificcvfinal\fi
\begin{document}

%%%%%%%%% TITLE
\title{NLNL: Negative Learning for Noisy Labels}

\author{Youngdong Kim ~~~~~~~~~~~~~ Junho Yim ~~~~~~~~~~~~~ Juseung Yun ~~~~~~~~~~~~~ Junmo Kim\\
School of Electrical Engineering, KAIST, South Korea \\
% Institution1 address\\
{\tt\small \{ydkim1293, junho.yim, st24hour, junmo.kim\}@kaist.ac.kr}
% For a paper whose authors are all at the same institution,
% omit the following lines up until the closing ``}''.
% Additional authors and addresses can be added with ``\and'',
% just like the second author.
% To save space, use either the email address or home page, not both
% \and
% Second Author\\
% Institution2\\
% First line of institution2 address\\
% {\tt\small secondauthor@i2.org}
}

\maketitle
\thispagestyle{empty}

%%%%%%%%% ABSTRACT
\begin{abstract}
Convolutional Neural Networks (CNNs) provide excellent performance when used for image classification. The classical method of training CNNs is by labeling images in a supervised manner as in ``input image belongs to this label'' (Positive Learning; PL), which is a fast and accurate method if the labels are assigned correctly to all images. However, if inaccurate labels, or noisy labels, exist, training with PL will provide wrong information, thus severely degrading performance. To address this issue, we start with an indirect learning method called Negative Learning (NL), in which the CNNs are trained using a complementary label as in ``input image does not belong to this complementary label.'' Because the chances of selecting a true label as a complementary label are low, NL decreases the risk of providing incorrect information. Furthermore, to improve convergence, we extend our method by adopting PL selectively, termed as \textit{Selective Negative Learning and Positive Learning} (SelNLPL). PL is used selectively to train upon expected-to-be-clean data, whose choices become possible as NL progresses, thus resulting in superior performance of filtering out noisy data. With simple semi-supervised training technique, our method achieves state-of-the-art accuracy for noisy data classification, proving the superiority of SelNLPL's noisy data filtering ability.
\end{abstract}
%-------------------------------------------------------------------------

\section{Introduction}
\label{sec:Introduction}

Convolutional Neural Networks (CNNs) have improved the performance of image classification significantly~\cite{krizhevsky2012imagenet, he2016deep, szegedy2015going, huang2017densely, han2017deep, zagoruyko2016wide}. For this supervised task, huge dataset composed of images and their corresponding labels is required for training CNNs. CNNs are powerful tools for classifying images if the corresponding labels are correct. However, accurately labeling a large number of images is daunting and time-consuming, occasionally yielding mismatched labeling. When the CNNs are trained with noisy data, it can overfit to such a dataset, resulting in poor classification performance. Therefore, training CNNs properly with noisy data is of great practical importance. Many approaches address this problem by applying a number of techniques and regularization terms along with \textit{Positive Learning} (PL), a typical supervised learning method for training CNNs that ``input image belongs to this label''~\cite{han2018co,ding2018semi,wang2018iterative,lee2017cleannet,ghosh2017robust,zhang2018generalized,reed2014training,tanaka2018joint,ma2018dimensionality,veit2017learning,li2017learning}. However, when the CNN is trained with images and mismatched labels, wrong information is being provided to the CNN.

\begin{figure}
\begin{center}
\includegraphics[width=8cm]{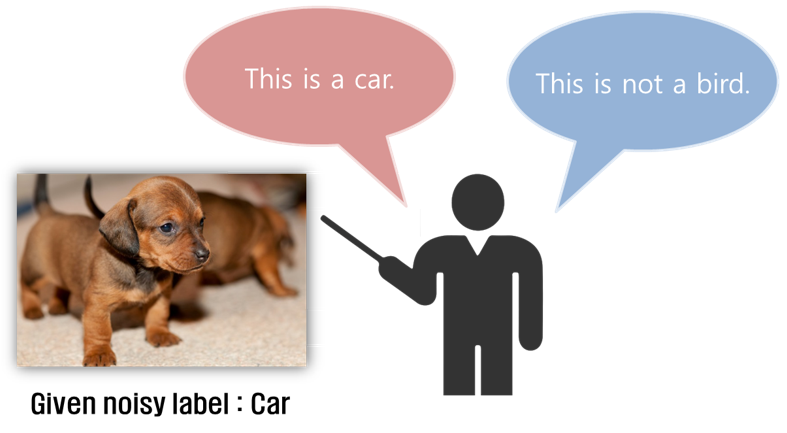}
\end{center}
\caption{Conceptual comparison between \textit{Positive Learning} (PL) and \textit{Negative Learning} (NL). Regarding noisy data, while PL provides CNN the wrong information (red balloon), with a higher chance, NL can provide CNN the correct information (blue balloon) because a dog is clearly not a bird.}
\label{fig:Overview}
\vspace{-3mm}
\end{figure}
% \begin{figure}
% \begin{center}
% \includegraphics[width=8cm]{Image/NLNLmain}
% \end{center}
% \caption{Conceptual comparison between \textit{Positive Learning} (PL) and \textit{Negative Learning} (NL). Regarding noisy data, while PL provides CNN the wrong information (red balloon), with a higher chance, NL can provide CNN the correct information (blue balloon) because a dog is clearly not a bird.}
% \label{fig:Overview}
% \vspace{-3mm}
% \end{figure}

To overcome this issue, we suggest \textit{Negative Learning} (NL), an indirect learning method for training CNN that ``input image does not belong to this complementary label.'' NL does not provide wrong information as frequently as PL (Figure~\ref{fig:Overview}). For example, when training CNN with noisy CIFAR10 using PL, if the CNN receives an image of a dog and the label ``car'', the CNN will be trained to acknowledge that this image is a car. In this case, the CNN is trained with wrong information. However, with NL, the CNN will be randomly provided with a complementary label other than ``car,'' for example, ``bird.'' Training CNN to acknowledge that this image is not a bird is in some way an act of providing CNN the right information because a dog is clearly not a bird. In this manner, noisy data can contribute to training CNN by providing the ``right'' information with a high chance of not selecting a true label as a complementary label, whereas zero chance is provided in PL. Our study demonstrates the effectiveness of NL as it prevents CNN from overfitting to noisy data.

Furthermore, utilizing NL training method, we propose \textit{Selective Negative Learning and Positive Learning} (SelNLPL), which combines PL and NL to take full advantage of both methods for better training with noisy data. Although PL is unsuitable for noisy data, it is still a fast and accurate method for clean data. Therefore, after training CNN with NL, PL begins to train CNN selectively using only training data of high classification confidence. Through this process, SelNLPL widens the gap between the confidences of clean data and noisy data, resulting in excellent performance for filtering noisy data from training data. 

Subsequently, by discarding labels of filtered noisy data and treating them as unlabeled data, we utilize semi-supervised learning for noisy data classification. Based on the superior filtering ability of SelNLPL, we demonstrate that state-of-the-art performance on noisy data classification can be achieved with a simple semi-supervised learning method. Although this is not the first time that noisy data classification has been addressed by filtering noisy data~\cite{ding2018semi,han2018co,northcutt2017learning}, the filtering results have not been promising owing to the use of PL for noisy data.

The main contributions of this paper are as follows:

\begin{description}[font=$\bullet$\scshape\bfseries, leftmargin=4mm, topsep=1mm, noitemsep]
  \item We apply the concept of \textit{Negative Learning} to the problem of noisy data classification. We prove its applicability by demonstrating that it prevents the CNN from overfitting to noisy data.
  \item Utilizing the proposed NL, we introduce a new framework, called SelNLPL, for filtering out noisy data from training data. Following NL, by selectively applying PL only to training data of high confidence, we can achieve accurate filtering of noisy data.
  \item We achieved state-of-the-art noisy data classification results with relatively simple semi-supervised learning based on the superior noisy data filtering achieved by SelNLPL.
  \item Our method does not require any prior knowledge of the type or number of noisy data points. It does not require any tuning of hyper-parameters that depend on prior knowledge, making our method applicable in real life.
\end{description}

The remainder of this paper is organized as follows: Section~\ref{sec:Method} describes the overall process of our method with detailed explanations of each step. Section~\ref{sec:Filtering ability} demonstrates the superior filtering ability of SelNLPL. Section~\ref{sec:Experiments} describes the experiments for evaluating our method, and Section~\ref{sec:Analysis} describes the experiments to further analyze our method. Finally, we conclude the paper in Section~\ref{sec:Conclusion}.
%-------------------------------------------------------------------------
\section{Related works}
\label{sec:Related works}

\noindent\textbf{Noisy label learning} Recently, numerous methods have been proposed for learning with noisy labels. Herein, we briefly review the relevant studies.

Some methods attempted to create noise-robust losses~\cite{brooks2011support,van2015learning,masnadi2009design,ghosh2015making,ghosh2017robust,zhang2018generalized}. Ghosh \etal~\cite{ghosh2015making,ghosh2017robust} demonstrated theoretically that the mean absolute error (MAE) is robust to noisy labels; however, the MAE can degrade the accuracy when adopted in neural networks. Zhang \etal~\cite{zhang2018generalized} proposed a generalized cross entropy loss that not only shows robustness to label noise but also performs well on deep neural networks. 

In other studies, each training sample is re-weighted differently depending on the reliability of the given label ~\cite{jiang2017mentornet,ren2018learning,lee2017cleannet}. 
~\cite{jiang2017mentornet,ren2018learning} used a meta-learning algorithm that learns the optimal weights for each sample. However, both these methods require a certain amount of clean data, which is difficult to obtain in many cases. 
CleanNet~\cite{lee2017cleannet} is also limited because it requires a label that is verified as correct. 

Some approaches use correction methods. Loss correction~\cite{patrini2017making,vahdat2017toward,xiao2015learning,hendrycks2018using} methods assume that the noise transition matrix is already known or that some clean data are obtainable to calculate the noise transition matrix.~\cite{sukhbaatar2014training, jindal2016learning, goldberger2016training} modeled the noise transition matrix by adding an additional layer. Several other approaches attempted to correct the label directly~\cite{veit2017learning, li2017learning}. However, in these methods, clean data is required to train the label cleaning network and teacher network. Additional label-cleaning methods exist, that gradually change the data label to the prediction value of the network~\cite{reed2014training,tanaka2018joint,ma2018dimensionality}.

Other approaches include jointly modeling labels and worker quality~\cite{khetan2017learning}, creating a robust method to learn in open-set noisy label situations~\cite{wang2018iterative} and attempting to prune the correct samples~\cite{han2018co,ding2018semi,northcutt2017learning}. Ding \etal~\cite{ding2018semi} suggested pruning the correct samples based on softmax outputs. Samples deemed unreliable are trained in a semi-supervised manner rather than by using label information. 

Our method utilizes both pruning of the correct sample and the label correction method. The existing pruning and label cleaning methods~\cite{han2018co,ding2018semi,reed2014training,tanaka2018joint,ma2018dimensionality} use a network trained directly with a given noisy label; thus overfitting to a noisy label can occur even if the pruning or cleaning process is performed. Meanwhile, we use NL method, which indirectly uses noisy labels, thereby avoiding the problem of memorizing the noisy label and exhibiting remarkable performance in filtering only noisy samples.

\noindent\textbf{Using complementary labels}
This is not the first time that complementary labels have been used. Previous studies~\cite{ishida2017learning,xiyueccv2018} focused on a classification task in which complementary labels are given. However, unlike the complementary label classification task, we generate complementary labels from given noisy labels and use them for NL.
%-------------------------------------------------------------------------
\begin{figure}
\begin{center}
\setlength{\tabcolsep}{2pt} % Default value: 6pt
\begin{tabular}{cc}
\includegraphics[width=4cm]{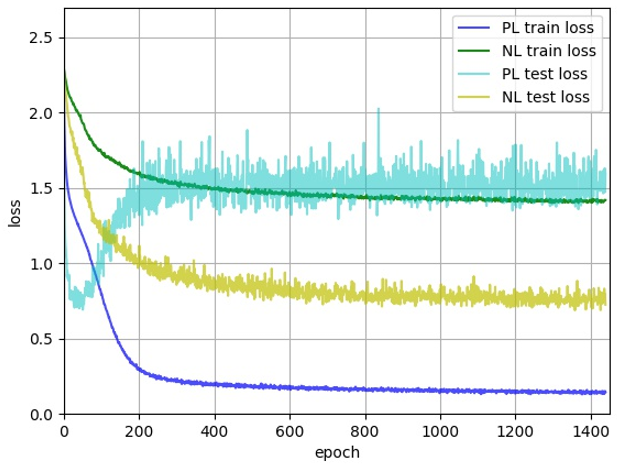}\vspace{-1pt} &\includegraphics[width=4cm]{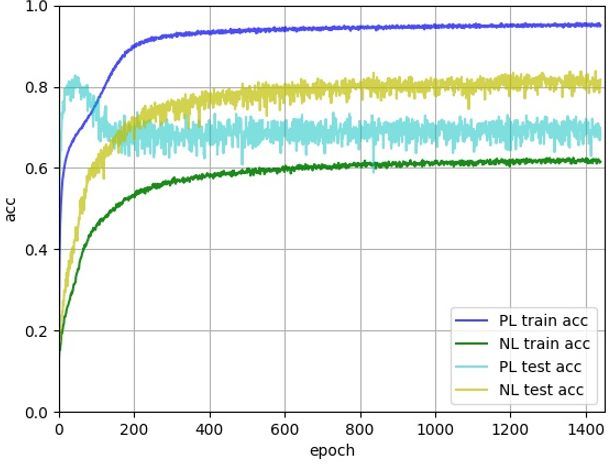}\vspace{-1pt}\\
\vspace{3mm}
\hspace{2.5mm}(a) & \hspace{5mm}(b)
\vspace{-3mm}
\end{tabular}
\end{center}
\caption{Comparison between PL and NL. (a): Loss graph of PL and NL. (b): Accuracy graph of PL and NL. }
\label{fig:PL_NL_graph}
\vspace{-5mm}
\end{figure}

\section{Method}
\label{sec:Method}
This section describes our overall method for noisy data classification. Section~\ref{sec:Negative learning} describes the concept and implementation of NL, demonstrating that it is more suitable for training with noisy data compared to PL. Section~\ref{sec:Selective NL}, and~\ref{sec:Selective PL} respectively introduce \textit{Selective Negative Learning} (SelNL) and \textit{Selective Positive Learning} (SelPL), which are the subsequent steps after NL to further make CNN train better with noisy training data and simultaneously prevent overfitting. The combination of all these methods is called \textit{Selective Negative Learning and Positive Learning} (SelNLPL), which demonstrates excellent performance for filtering noisy data from training data (Section~\ref{sec:Selective NL and PL}). Finally, semi-supervised learning is performed for noisy data classification, utilizing the filtering ability of SelNLPL (Section~\ref{sec:Semi-supervised learning}).

\subsection{Negative Learning}
\label{sec:Negative learning}
As mentioned in Section~\ref{sec:Introduction}, typical method of training CNNs for image classification with given image data and the corresponding labels is PL. It is a method of training  CNNs ``input image belongs to this label.'' In contrast, with NL, the CNNs are trained that ``input image does not belong to this complementary label.''

\begin{algorithm}
\caption{Complementary label generation}\label{alg:Complementary label generation}
\textbf{Input:} Training label $y \in \mathcal{Y}$
\begin{algorithmic}

\While{iteration}
\State $\overline{y}$ = Randomly select from $\{1,...,C\}\setminus\{y\}$
\EndWhile

\end{algorithmic}
\textbf{Output:} Complementary label $\overline{y}$
\end{algorithm}
\vspace{-3mm}

We consider the problem of c-class classification. Let \( \boldsymbol{x} \in \mathcal{X} \) be an input, \( y, \overline{y} \in \mathcal{Y} = \{1,...,c\}\) be its label and complementary label, respectively, and \( \boldsymbol{y}, \boldsymbol{\overline{y}} \in \{0,1\}^{c}\) be their one-hot vector. Suppose the CNN $f(x;\theta)$ maps the input space to the c-dimensional score space \( f : \mathcal{X} \rightarrow \mathbb{R}^c \), where $\theta$ is the set of network parameters. 
If $f$ passes through the softmax function, the output can be interpreted as a probability \(\boldsymbol{p} \in \Delta^{c-1} \), where \(\Delta^{c-1}\) denotes the c-dimensional simplex.
When training with PL, the cross entropy loss function of the network \(f\) becomes:
\vspace{-2mm}
\begin{equation}
\label{eq:PL}
\mathcal{L}(f,y) = -\sum_{k=1}^c \boldsymbol{y}_k \log \boldsymbol{p}_k
\vspace{-2mm}
\end{equation}
where $\boldsymbol{p}_k$ denotes the $k^{th}$ element of $\boldsymbol{p}$. Eq.~\ref{eq:PL} is suitable for optimizing the probability value corresponding to the given label as 1 ($\boldsymbol{p}_y \rightarrow 1$), satisfying the purpose of PL. However, NL differs from PL as it optimizes the output probability corresponding to the complementary label to be far from 1 (to reach 0 in the end ($\boldsymbol{p}_{\overline{y}}\rightarrow 0$)). Therefore, we propose a loss function as follows:
\vspace{-2mm}
\begin{equation}
\label{eq:NL}
\mathcal{L}(f,\overline{y}) = -\sum_{k=1}^c \boldsymbol{\overline{y}}_k \log(1-\boldsymbol{p}_k)
\vspace{-2mm}
\end{equation}
This complementary label is completely random in that it is selected randomly from the labels of all classes except for the given label \( y \) for every iteration during training (Algorithm~\ref{alg:Complementary label generation}). 
Eq.~\ref{eq:NL} enables the probability value of the complementary label to be optimized as zero, resulting in an increase in the probability values of other classes,
% . The other leftover probability values are subsequently increased, 
meeting the purpose of NL.

\begin{figure*}[t]

\begin{center}
\begin{tabular}{cccc}
\includegraphics[width=4.05cm]{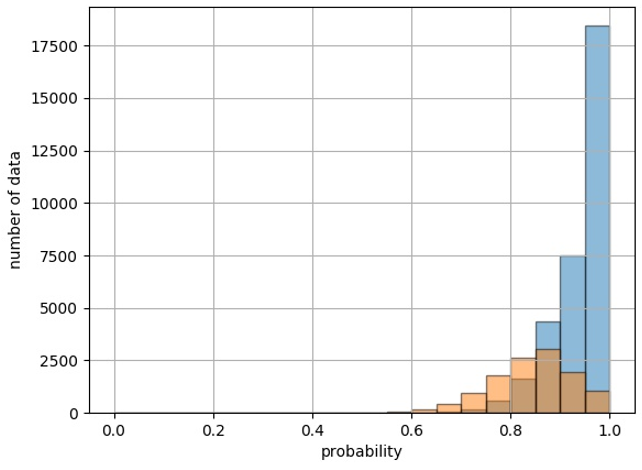} &\includegraphics[width=4cm]{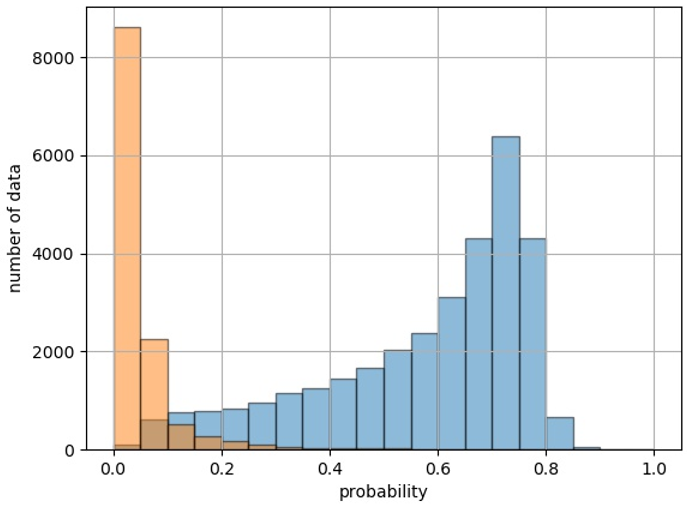} & \includegraphics[width=4.05cm]{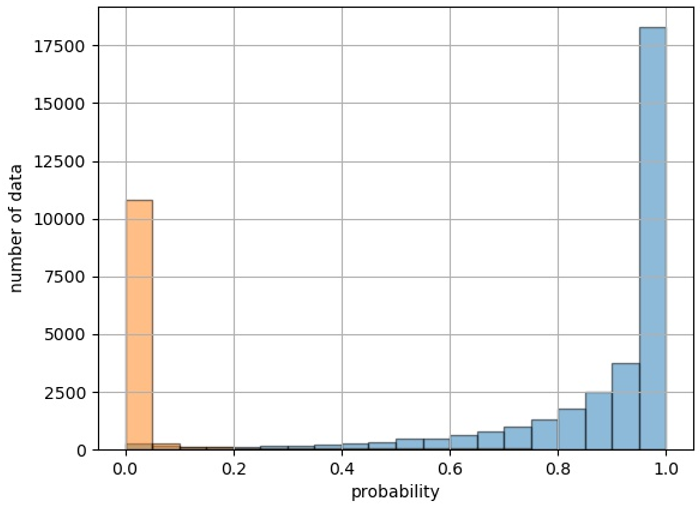} & \includegraphics[width=4.05cm]{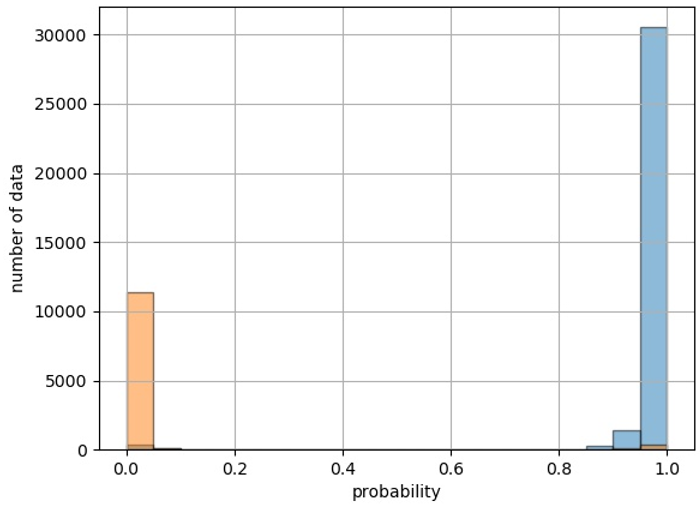} \\
(a) & (b) & (c) & (d)\\
\end{tabular}
\end{center}

\caption{Histogram showing the distribution of CIFAR10 training data with 30\% \textit{symm-inc} noise, according to probability $\boldsymbol{p}_y$ (confidence). Blue indicates clean data, whereas orange indicates noisy data. (a): PL. (b): NL. (c): NL$\rightarrow$SelNL. (d): NL$\rightarrow$SelNL$\rightarrow$SelPL (SelNLPL).}
\label{fig:SelPLNL_results}
\vspace{-5mm}
\end{figure*}

A distinct comparison between PL and NL is shown in Figure~\ref{fig:PL_NL_graph}. The CNN was trained with either PL or NL on CIFAR10 corrupted with 30\% \textit{symm-inc} noise. The types of noise used in our paper are explained in Section~\ref{sec:Experiments}. Note that, while the CNN is trained with either PL (Eq.~\ref{eq:PL}) or NL (Eq.~\ref{eq:NL}), all losses shown in the graph (Figure~\ref{fig:PL_NL_graph} (a)) are computed using Eq.~\ref{eq:PL}. With PL, the test loss drops and the test accuracy increases in the early stage. However, it eventually causes the CNN to overfit to noisy training data, resulting in poor performance on the clean test data. In contrast, NL is shown to train the CNN without overfitting to noisy training data, as gradual decrease in test loss and an increase in test accuracy are observed. Figure~\ref{fig:SelPLNL_results} (a) and (b) respectively show the histogram of the training data after PL and NL. While the confidence of both clean and noisy data increased with PL, the confidence of noisy data is much lower than that of clean data with NL, again indicating the capability of NL to prevent the CNN from overfitting to noisy data.

\begin{figure}[t]
\begin{center}
\begin{tabular}{c}
\includegraphics[width=5cm]{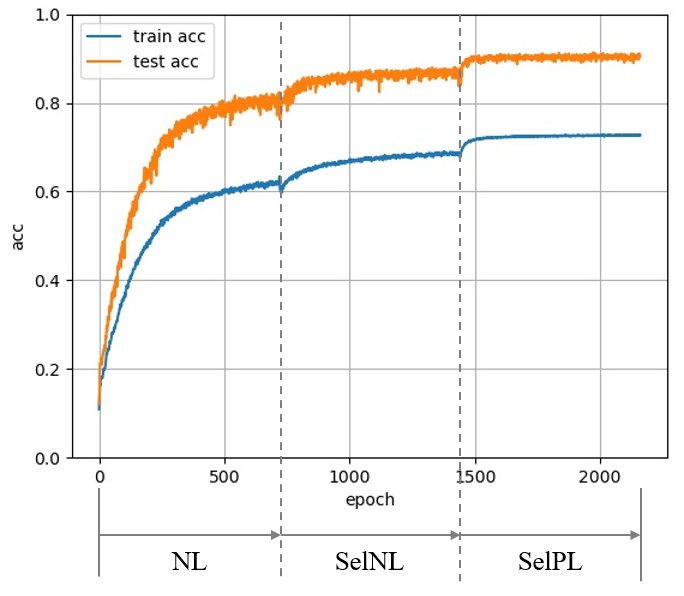} \\
\end{tabular}
\end{center}
\caption{Accuracy graph of SelNLPL. Training is performed sequentially with NL, SelNL, and SelPL.}
\label{fig:SelPLNL_acc_graph}
\vspace{-5mm}
\end{figure}
%-------------------------------------------------------------------------
\subsection{Selective NL}
\label{sec:Selective NL}
As mentioned in Section~\ref{sec:Negative learning}, NL can prevent the CNN from overfitting to noisy data, as shown by its low confidence values (Figure~\ref{fig:SelPLNL_results} (b)). As the next step, we introduce SelNL to improve convergence after NL. After training with NL, SelNL trains the CNN only with the data having confidence over $\frac{1}{c}$. After thresholding, the data involved in training tends to be less noisy than before, thus improving the convergence of the CNN efficiently. Figure~\ref{fig:SelPLNL_results} (c) shows the result of SelNL after NL.
%-------------------------------------------------------------------------
\begin{figure*}[t]
\begin{center}
\includegraphics[width=11cm]{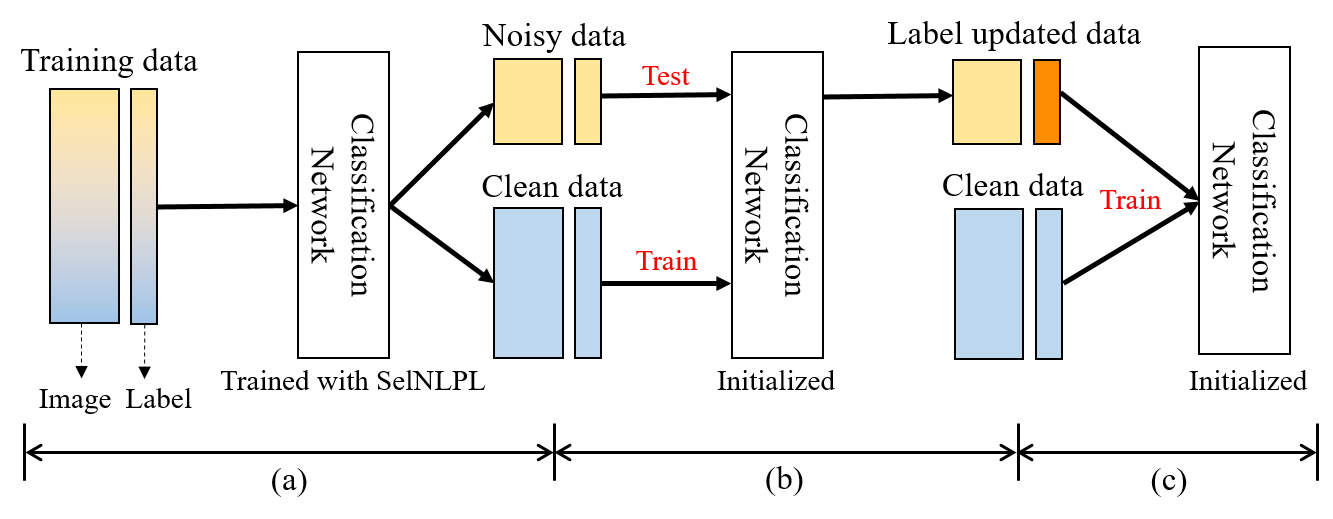}\vspace{-5pt}
\end{center}
\caption{Pseudo labeling for semi-supervised learning. (a): Division of training data into either clean or noisy data with CNN trained with SelNLPL. (b): Training initialized CNN with clean data from (a), then noisy data's label is updated following the output of CNN trained with clean data. (c): Clean data and label-updated noisy data are both used for training initialized CNN in the final step.}
\label{fig:Semi_supervised_learning}
\vspace{-5mm}
\end{figure*}

\begin{figure}[h]
\begin{center}
\begin{tabular}{c}
\includegraphics[width=5.5cm]{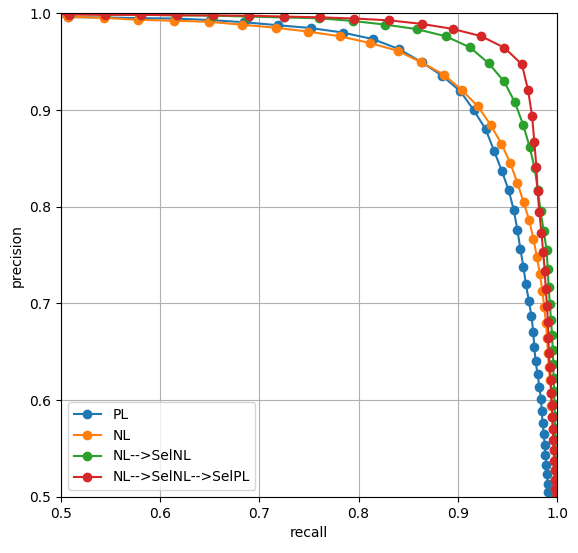}\vspace{-5pt} \\
\end{tabular}
\end{center}
\caption{Precision-Recall curve when filtering noisy data. Each curve represents the filtering performance of PL, NL, NL$\rightarrow$SelNL, and NL$\rightarrow$SelNL$\rightarrow$SelPL (SelNLPL).}
\label{fig:Precision-Recall Curve}
\vspace{0mm}
\end{figure}

\setlength{\textfloatsep}{2mm}
\begin{algorithm}[t]
\caption{Overall process of SelNLPL}\label{alg:Overall process of SelNLPL}
\textbf{Input:} Training data $(\boldsymbol{x},y) \in (\mathcal{X},\mathcal{Y})$, network $f(\boldsymbol{x};\boldsymbol{\theta)}$, total epoch $T$
\begin{algorithmic}%[1]

\For{$i \gets 1$ to $T$}\Comment{NL}
\State Batch $\gets$ Sample $\boldsymbol{x}$ 
\State Update $f$ by minimizing Eq.~\ref{eq:NL}
\EndFor

\For{$i \gets 1$ to $T$}\Comment{SelNL}
\State Batch $\gets$ Sample $\boldsymbol{x}$ if $\boldsymbol{p}_y > 1/$c$ $
\State Update $f$ by minimizing Eq.~\ref{eq:NL} 
\EndFor

\For{$i \gets 1$ to $T$}\Comment{SelPL}
\State Batch $\gets$ Sample $\boldsymbol{x}$ if $\boldsymbol{p}_y > \gamma$
\State Update $f$ by minimizing Eq.~\ref{eq:PL} 
\EndFor

\end{algorithmic}
\textbf{Output: } Network $f(\boldsymbol{x};\boldsymbol{\theta)}$
\end{algorithm}

\subsection{Selective PL}
\label{sec:Selective PL}
NL can be a better learning method when noisy data is involved. However, if the training data is verified to have clean labels, PL is a faster and more accurate method than NL. After training with NL and SelNL, the confidences of clean and noisy data are separated by a large margin (Figure~\ref{fig:SelPLNL_results} (c)). SelPL trains CNN only with data having confidence over $\gamma$, assuming that such data is clean data. In this study, we set $\gamma$ to 0.5. Figure~\ref{fig:SelPLNL_results} (d) shows the result of SelPL after Figure~\ref{fig:SelPLNL_results} (c), exhibiting high confidence value near 1 for almost all clean data.
%-------------------------------------------------------------------------
\subsection{Selective NL and PL}
\label{sec:Selective NL and PL}

To summarize, the combination of NL, SelNL, and SelPL is called SelNLPL. The overall process of SelNLPL is shown in Algorithm~\ref{alg:Overall process of SelNLPL}. Figure~\ref{fig:SelPLNL_acc_graph} displays the performance change for each step. It clearly indicates the enhancement in performance when each step is applied, thus demonstrating the significance of each step in SelNLPL. It is proven in Figure~\ref{fig:SelPLNL_acc_graph} that each step of SelNLPL contributes to convergence, while simultaneously preventing overfitting to noisy data, resulting in a higher test accuracy than training accuracy throughout the training process.

As shown in Figure~\ref{fig:SelPLNL_results} (d), the overall confidence of clean data and noisy data is separated with a large margin. This implies that SelNLPL can be used for filtering noisy data from training data. This area is further analyzed in Section~\ref{sec:Filtering ability}.
%-------------------------------------------------------------------------
\subsection{Semi-supervised learning}
\label{sec:Semi-supervised learning}
With the filtering ability of SelNLPL, the semi-supervised learning method can be applied to clean data and filtered noisy data, discarding the labels of filtered noisy data. For semi-supervised learning, we apply the pseudo labeling method~\cite{lee2013pseudo}. Figure~\ref{fig:Semi_supervised_learning} shows the overall process of pseudo labeling. Firstly, the training data is divided into clean data and noisy data by using the CNN that is trained with SelNLPL (Figure~\ref{fig:Semi_supervised_learning} (a)). 
Next, in Figure~\ref{fig:Semi_supervised_learning} (b), the initialized CNN is trained with clean data obtained from SelNLPL. Then, the noisy data's label is updated with the output of the CNN in Figure~\ref{fig:Semi_supervised_learning} (b). Here, we used the soft label as the updated label, similar to~\cite{tanaka2018joint}. The typical label for image classification is in the form of a one-hot vector, whereas the soft label is simply the output of trained CNN. It was shown that soft labels are better when updating labels~\cite{tanaka2018joint}. Finally, clean data and label-updated noisy data are used to train the initialized CNN (Figure~\ref{fig:Semi_supervised_learning} (c)). This resulted in state-of-the-accuracy, proving the high filtering ability of SelNLPL. The results are shown in Section~\ref{sec:Experiments}.

\begin{table}
\footnotesize
\begin{center}
\begin{tabular}{cc|ccc}
\hline
\multirow{2}{*}{} & \multirow{2}{*}{} & \multirow{2}{*}{Estimated noise (\%)} & \multirow{2}{*}{Recall} & \multirow{2}{*}{Precision} \\
& & & & \\
\hline
\multirow{3}{*}{Noise (\%)} & \multicolumn{1}{|c|}{9} & 10.25 & 92.87 & 85.20 \\
 & \multicolumn{1}{|c|}{27} & 27.60 & 96.28 & 94.01 \\
 & \multicolumn{1}{|c|}{45} & 45.15 & 95.80 & 95.38 \\
\hline
\end{tabular}
\end{center}
\caption{Results for filtering noisy data from CIFAR10 with 10\%, 30\%, and 50\% \textit{symm-inc} noise. Owing to the characteristics of \textit{symm-inc} noise (Section~\ref{sec:Experimental settings}), the actual noise in each case is 9\%, 27\%, and 45\%, respectively.}
\label{tab:Filtering result table}
\vspace{0mm}
\end{table}
%-------------------------------------------------------------------------
\section{Filtering ability}
\label{sec:Filtering ability}
It is mentioned in Section~\ref{sec:Selective NL and PL} that SelNLPL is effective for filtering noisy data from training data. In this section, we explain the filtering process of SelNLPL further.

When training CNN with SelNLPL, data with confidence exceeding $\gamma$ are assumed to be clean. Following this approach, we filter out data that were not trained with PL as noisy data. Table~\ref{tab:Filtering result table} summarizes the filtering results for SelNLPL on CIFAR10 with various noise ratios (\textit{symm-inc}). The estimated noise ratio refers to the amount of data that was not trained with PL. Recall and Precision are measures for the quality of noisy data filtering. It shows that the estimated noise ratio almost matches the actual noise ratio by 88\% to 99\%. Furthermore, Table~\ref{tab:Filtering result table} shows our method of filtering noisy data resulted in high recall and precision values, indicating that our method filtered out most of the pure noisy data from the training data. This implies that even if the amount of noise mixed in the training data is unknown, which is normal in practical situations, the amount of noise can be estimated with SelNLPL, which is a huge advantage because it can be used as an indicator of the training data quality. 

Figure~\ref{fig:Precision-Recall Curve} compares the overall filtering ability of the proposed methods. The curve for PL is obtained when the CNN is trained with PL before overfitting to noisy data. The curve indicates that each step of SelNLPL contributes to an increase in filtering performance, surpassing that of PL.

To summarize, SelNLPL exhibits excellent results for filtering noisy data from training data, as shown in Figure~\ref{fig:Precision-Recall Curve}. Additionally, the estimated noise ratio from SelNLPL almost matches that of the actual noise ratio; hence, can be used to indicate training data quality in practical situations where the actual noise ratio is not available. 
%-------------------------------------------------------------------------
\section{Experiments}
\label{sec:Experiments}
% Overall accuracy
In this section, we describe the experiments performed to evaluate our method. The results of pseudo labeling after SelNLPL are compared to those of other existing methods for noisy data classification. To show that our method generalizes to various environments, we follow each different experimental settings of other baseline methods, which vary in terms of CNN architecture, dataset, and type of noise in the training data. (Table~\ref{tab:Generalized},~\ref{tab:Joint},~\ref{tab:Dimensionality},~\ref{tab:Semi})

\subsection{Experimental settings}
\label{sec:Experimental settings}
We conducted our experiments following experimental settings of four different baselines. The experimental results of each baseline are reported, and we added our results to each of those result tables for comparison (Table~\ref{tab:Generalized},~\ref{tab:Joint},~\ref{tab:Dimensionality},~\ref{tab:Semi}).
Table~\ref{tab:Experimental settings} summarizes different experimental settings for each baseline. Datasets we used are CIFAR10, CIFAR100~\cite{cifar-10}, FashionMNIST~\cite{xiao2017fashion}, and MNIST~\cite{lecun-mnisthandwrittendigit-2010}. 

We applied three different types of noise following the baseline methods. 

\textbf{Symmetric noise} The basic idea of symmetric noise is to randomly select a label with equal probabilities among the classes. In this experiment, two symmetric noises were used: \textit{symm-inc} noise and \textit{symm-exc} noise. \textit{Symm-inc} noise is created by randomly selecting the label from all classes, including the ground truth label, whereas \textit{symm-exc} noise flips the ground truth label to one of the other class labels, thus excluding the ground truth label. \textit{Symm-inc} noise is used in Table~\ref{tab:Joint}, and \textit{symm-exc} noise is used in Table~\ref{tab:Generalized},~\ref{tab:Dimensionality},~\ref{tab:Semi}.

\textbf{Asymmetric noise} As described by Patrini \etal~\cite{patrini2017making}, this noise mimics some of the structures of real errors for similar classes. For CIFAR10, asymmetric (\textit{asymm}) noise was generated by mapping TRUCK $\rightarrow$ AUTOMOBILE, BIRD $\rightarrow$ PLANE, DEER $\rightarrow$ HORSE, and CAT $\leftrightarrow$ DOG. For FashionMNIST, BOOT $\rightarrow$ SNEAKER, SNEAKER $\rightarrow$ SANDALS, PULLOVER $\rightarrow$ SHIRT, and COAT $\leftrightarrow$ DRESS were mapped, following ~\cite{zhang2018generalized}. For MNIST, 2 $\rightarrow$ 7, 3 $\rightarrow$ 8, 7 $\rightarrow$ 1, and 5 $\leftrightarrow$ 6 were mapped, following ~\cite{patrini2017making}. For CIFAR100, the noise flipped each class into the next, circularly within super-classes, following~\cite{patrini2017making}.

For optimization, we used stochastic gradient descent (SGD) with a momentum of  0.9, weight decay of $10^{-4}$, and batch size of 128. For NL, SelNL, and SelPL, each of them trained the CNN for 720 epochs. Except for MNIST, learning rate was unified across all datasets and CNN architectures. 
Learning rates for NL, SelNL, and SelPL were set to 0.02, 0.02, and 0.1, respectively. For pseudo labeling, for each step (Figure~\ref{fig:Semi_supervised_learning} (b), (c)), learning rate was scheduled to start from 0.1 and was divided by 10 at 192, 288 epochs (480 epochs total). As an exception, learning rates for NL and SelNL were set to 0.1 when using MNIST.

\begin{table}
\footnotesize
\begin{center}
\begin{tabular}{cc|cccc}
\hline
 & & Table~\ref{tab:Generalized} & Table~\ref{tab:Joint} & Table~\ref{tab:Dimensionality} & Table~\ref{tab:Semi} \\
\hline
\multicolumn{2}{c|}{Validation split} & \checkmark & \checkmark & x & \checkmark \\
\hline
\multicolumn{1}{c|}{} & \textit{symm inc} & x & \checkmark & x & x \\
\multicolumn{1}{c|}{noise} & \textit{symm exc} & \checkmark & x & \checkmark & \checkmark \\
\multicolumn{1}{c|}{} & \textit{asymm} & \checkmark & \checkmark & x & \checkmark \\
\hline
\end{tabular}
\end{center}
\caption{Experimental settings (dataset preparation and types of noise) used for Table~\ref{tab:Generalized},~\ref{tab:Joint},~\ref{tab:Dimensionality},~\ref{tab:Semi}.}
\label{tab:Experimental settings}
\vspace{0mm}
\end{table}

\begin{table*}
\footnotesize
\begin{center}
\begin{tabular}{c|c|c|cccc|cccc}
\hline
\multirow{2}{*}{Datasets} & \multirow{2}{*}{Model} & \multirow{2}{*}{Methods} & \multicolumn{4}{c|}{\textit{Symm}} & \multicolumn{4}{c}{\textit{Asymm}} \\
 & & & 20 & 40 & 60 & 80 & 10 & 20 & 30 & 40 \\
\hline
\multirow{7}{*}{FashionMNIST} & \multirow{7}{*}{ResNet18} & CE  & 93.24 & 92.09 & 90.29 & 86.20 & 94.06 & 93.72 & 92.72 & 89.82\\
 & & MAE~\cite{ghosh2017robust} & 80.39 & 79.30 & 82.41 & 74.73 & 74.03 & 63.03 & 58.14 & 56.04\\
 & & Forward $T$~\cite{patrini2017making} & 93.64 & 92.69 & 91.16 & 87.59 & 94.33 & 94.03 & 93.91 & 93.65\\
 & & Forward $\hat{T}$~\cite{patrini2017making} & 93.26 & 92.24 & 90.54 & 85.57 & 94.09 & 93.66 & 93.52 & 88.53\\
 & & $L_q$~\cite{zhang2018generalized} & 93.35 & 92.58 & 91.30 & 88.01 & 93.51 & 93.24 & 92.21 & 89.53\\
 & & Truncated $L_q$~\cite{zhang2018generalized} & 93.21 & 92.60 & 91.56 & 88.33 & 93.53 & 93.36 & 92.76 & 91.62\\
 & & Ours & \textbf{94.82} & \textbf{94.16} & \textbf{92.78} & - & \textbf{95.10} & \textbf{94.88} & \textbf{94.66} & \textbf{93.96} \\
\hline
\multirow{7}{*}{CIFAR10} & \multirow{7}{*}{ResNet34} & CE & 86.98 & 81.88 & 74.14 & 53.82 & 90.69 & 88.59 & 86.14 & 80.11\\
 & & MAE~\cite{ghosh2017robust} & 83.72 & 67.00 & 64.21 & 38.63 & 82.61 & 52.93 & 50.36 & 45.52\\
 & & Forward $T$~\cite{patrini2017making} & 88.63 & 85.07 & 79.12 & 64.30 & 91.32 & 90.35 & 89.25 & 88.12\\
 & & Forward $\hat{T}$~\cite{patrini2017making} & 87.99 & 83.25 & 74.96 & 54.64 & 90.52 & 89.09 & 86.79 & 83.55\\
 & & $L_q$~\cite{zhang2018generalized} & 89.83 & 87.13 & 82.54 & 64.07 & 90.91 & 89.33 & 85.45 & 76.74\\
 & & Truncated $L_q$~\cite{zhang2018generalized} & 89.70 & 87.62 & 82.70 & 67.92 & 90.43 & 89.45 & 87.10 & 82.28\\
%  & & Ours(0.01) & 93.58 & 91.51 & 85.98 & - & 93.5 & 90.84 & 86.14 & 79.26\\
 & & Ours & \textbf{94.23} & \textbf{92.43} & \textbf{88.32} & - & \textbf{94.57} & \textbf{93.35} & \textbf{91.80} & \textbf{89.86}\\
\hline
\multirow{7}{*}{CIFAR100} & \multirow{7}{*}{ResNet34} & CE & 58.72 & 48.20 & 37.41 & 18.10 & 66.54 & 59.20 & 51.40 & 42.74\\
 & & MAE~\cite{ghosh2017robust} & 15.80 & 9.03 & 7.74 & 3.76 & 13.38 & 11.50 & 8.91 & 8.20 \\
 & & Forward $T$~\cite{patrini2017making} & 63.16 & 54.65 & 44.62 & 24.83 & \textbf{71.05} & \textbf{71.08} & \textbf{70.76} & \textbf{70.82}\\
 & & Forward $\hat{T}$~\cite{patrini2017making} & 39.19 & 31.05 & 19.12 & 8.99 & 45.96 & 42.46 & 38.13 & 34.44\\
 & & $L_q$~\cite{zhang2018generalized} & 66.81 & 61.77 & 53.16 & 29.16 & 68.36 & 66.59 & 61.45 & 47.22\\
 & & Truncated $L_q$~\cite{zhang2018generalized} & 67.61 & 62.64 & 54.04 & 29.60 & 68.86 & 66.59 & 61.87 & 47.66\\
%  & & Ours(0.01) & 93.58 & 91.51 & 85.98 & - & 93.5 & 90.84 & 86.14 & 79.26\\
 & & Ours & \textbf{71.52} & \textbf{66.39} & \textbf{56.51} & - & 70.35 & 63.12 & 54.87 & 45.70\\
\hline
\end{tabular}
\end{center}
\caption{Comparison with results reported by Zhang \textit{et al.}~\cite{zhang2018generalized}}
\label{tab:Generalized}
\vspace{-2mm}
\end{table*}

\begin{table*}
\footnotesize
\begin{center}
\begin{tabular}{c|c|c|cccc|cccc}
\hline
\multirow{2}{*}{Datasets} & \multirow{2}{*}{Model} & \multirow{2}{*}{Methods} & \multicolumn{4}{c|}{\textit{Symm}} & \multicolumn{4}{c}{\textit{Asymm}} \\
 & & & 10 & 30 & 50 & 70 & 10 & 20 & 30 & 40 \\
\hline
\multirow{7}{*}{CIFAR10} & \multirow{7}{*}{Pre-ResNet32} & CE & 87.0 & 72.2 & 55.3 & 36.6 & 89.8 & 85.4 & 81.0 & 75.7\\
 & & Forward~\cite{patrini2017making} & - & - & - & - & 91.7 & 89.7 & 88.0 & 86.4\\
 & & CNN-CRF~\cite{vahdat2017toward} & - & - & - & - & 90.3 & 86.6 & 83.6 & 79.7\\
 & & Joint\hspace{0.3cm}~\cite{tanaka2018joint} & 92.9 & 91.5 & 89.8 & 86.0 & 93.2 & 92.8 & 92.4 & \textbf{91.7} \\
 & & Joint*\hspace{0.15cm}~\cite{tanaka2018joint} & - & - & - & - & 93.05 & 92.60 & 91.59 & 89.23 \\
 & & Joint**~\cite{tanaka2018joint} & - & - & - & - & 93.26 & 93.06 & 92.04 & 90.65 \\
 & & Ours & \textbf{94.25} & \textbf{93.42} & \textbf{91.45} & \textbf{86.13} & \textbf{94.12} & \textbf{93.44} & \textbf{92.56} & 90.99 \\
\hline
\end{tabular}
\end{center}
\caption{Comparison with results reported by Tanaka \textit{et al.}~\cite{tanaka2018joint}}
\label{tab:Joint}
\vspace{-5mm}
\end{table*}

\begin{table}
\footnotesize
\begin{center}
\begin{tabular}{c|c|c|ccc}
\hline
\multirow{2}{*}{Datasets} & \multirow{2}{*}{Model} & \multirow{2}{*}{Methods} & \multicolumn{3}{c}{\textit{Symm}} \\
 & & & 20 & 40 & 60\\
\hline
\multirow{7}{*}{MNIST} & \multirow{7}{*}{LeNet} & CE & 88.02 & 68.46 & 45.51\\
 & & Forward~\cite{patrini2017making} & 96.45 & 94.90 & 82.88\\
 & & Backward~\cite{patrini2017making} & 90.12 & 70.89 & 52.83\\
 & & Boot-hard~\cite{reed2014training} & 87.69 & 69.49 & 50.45\\
 & & Boot-soft~\cite{reed2014training} & 88.50 & 70.19 & 46.04\\
 & & D2L~\cite{ma2018dimensionality} & 98.84 & 98.49 & 94.73\\
 & & Ours & \textbf{99.35} & \textbf{99.27} & \textbf{98.91}\\
\hline
\end{tabular}
\end{center}
\caption{Comparison with results reported by Ma \textit{et al.}~\cite{ma2018dimensionality}}
\label{tab:Dimensionality}
\vspace{0mm}
\end{table}

\begin{table}
\footnotesize
\begin{center}
\begin{tabular}{p{10mm}|p{8mm}|c|c|cc}
\hline
\multirow{2}{*}{Datasets} & \multirow{2}{*}{Model} & \multirow{2}{*}{Methods} & \textit{Symm} & \multicolumn{2}{c}{\textit{Asymm}} \\
 & & & 20 & 20 & 60\\
\hline
\multirow{10}{*}{CIFAR10} & \multirow{10}{*}{\hspace{-1.5mm}ResNet14} & CE & 83.7 & 85.0 & 57.6\\
 & & Unhinged (BN)~\cite{van2015learning} & 84.1 & 83.8 & 52.1\\
 & & Sigmoid (BN)~\cite{ghosh2015making} & 66.6 & 71.8 & 57.0\\
 & & Savage~\cite{masnadi2009design} & 77.4 & 76.0 & 50.5\\
 & & Bootstrap soft~\cite{reed2014training} & 84.3 & 84.6 & 57.8\\
 & & Bootstrap hard~\cite{reed2014training} & 83.6 & 84.7 & 58.3\\
 & & Backward~\cite{patrini2017making} & 80.4 & 83.8 & 66.7\\
 & & Forward~\cite{patrini2017making} & 83.4 & 87.0 & 74.8\\
 & & Semi~\cite{ding2018semi} & 84.5 & 85.6 & \textbf{75.8}\\
 & & Ours & \textbf{89.85} & \textbf{90.1} & 74.44 \\
\hline
\multirow{10}{*}{MNIST} & \multirow{10}{*}{\hspace{2mm}FC2} & CE & 96.9 & 97.5 & 53.0\\
 & & Unhinged (BN)~\cite{van2015learning} & 96.9 & 97.0 & 71.2\\
 & & Sigmoid (BN)~\cite{ghosh2015making} & 93.1 & 96.7 & 71.4\\
 & & Savage~\cite{masnadi2009design} & 96.9 & 97.0 & 51.3\\
 & & Bootstrap soft~\cite{reed2014training} & 96.9 & 97.5 & 53.0\\
 & & Bootstrap hard~\cite{reed2014training} & 96.8 & 97.4 & 55.0\\
 & & Backward~\cite{patrini2017making} & 96.9 & 96.7 & 67.4\\
 & & Forward~\cite{patrini2017making} & 96.9 & 97.7 & 64.9\\
 & & Semi~\cite{ding2018semi} & 97.7 & 97.8 & \textbf{83.4}\\
 & & Ours & \textbf{97.96} & \textbf{97.97} & 79.48 \\
\hline
\end{tabular}
\end{center}
\caption{Comparison with results reported by Ding \textit{et al.}~\cite{ding2018semi}}
\label{tab:Semi}
\vspace{0mm}
\end{table}
% Table 간 간격 조절

\begin{figure*}
\begin{center}
\begin{tabular}{cccc} % 5.5
\includegraphics[width=4cm]{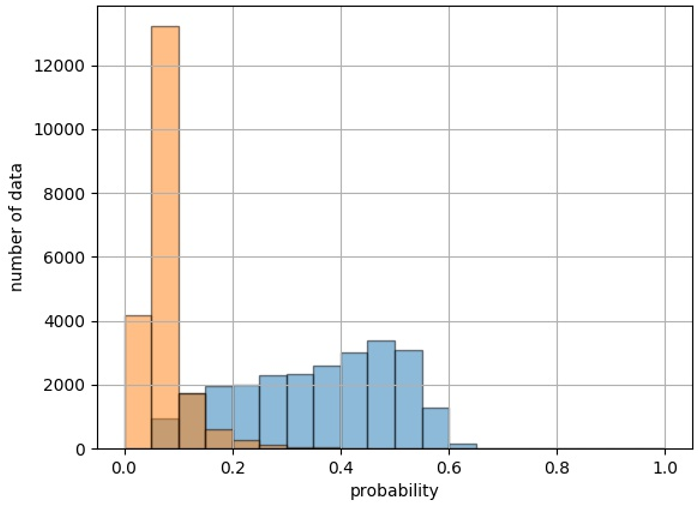} & \includegraphics[width=4cm]{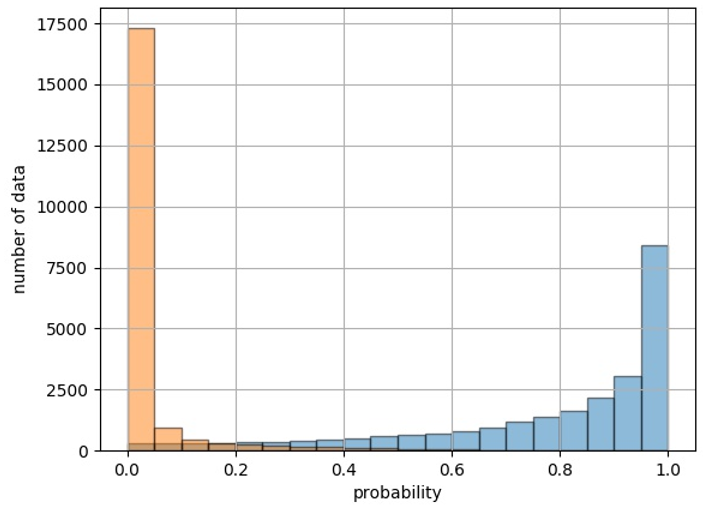} & \includegraphics[width=4cm]{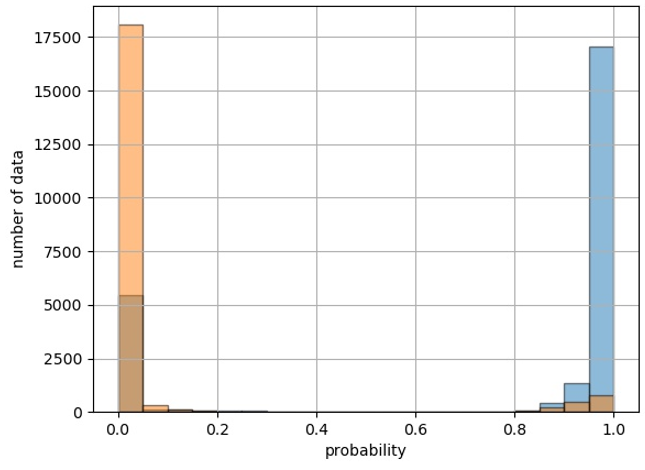} & \includegraphics[width=4cm]{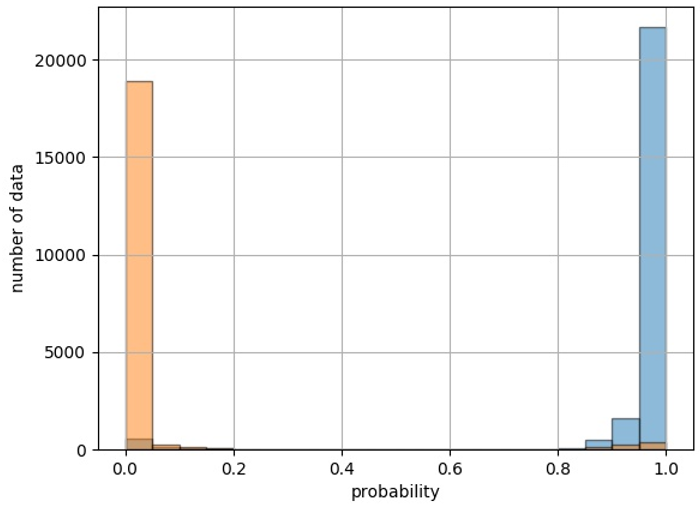} \\
(a) & (b) & (c) & (d) \\
\end{tabular}
\end{center}
\caption{Histogram showing the distribution of CIFAR10 training data with 50\% \textit{symm-inc} noise, according to probability (confidence). (a): NL. (b): NL$\rightarrow$SelNL. (c): NL$\rightarrow$SelPL. (d): NL$\rightarrow$SelNL$\rightarrow$SelPL (SelNLPL).}
\label{fig:SelPLNL_results-50}
\end{figure*}

\begin{table*}
\footnotesize
\begin{center}
\begin{tabular}{cc|cccc|cccc}
\hline 
  & & \multicolumn{4}{c}{30\% \textit{symm-inc} noise} & \multicolumn{4}{c}{50\% \textit{symm-inc} noise} \\
 \cline{3-10}
  & & Accuracy & Estimated noise (\%) & Recall & Precision & Accuracy & Estimated noise (\%) & Recall & Precision \\
 \hline
 \#1 & NL-SelNL-SelPL & 93.82 & 27.60 & 96.28 & 94.01 & 91.17 & 45.15 & 95.80 & 95.38 \\
 \#2 &  NL-SelNL & 92.44 (-1.38) & 33.76 & 98.47 & 78.62 & 89.36 (-1.81) & 52.75 & 98.56 & 83.99\\
 \#3 &  NL-SelPL & 93.41 (-0.41) & 28.17 & 97.04 & 92.84 & 72.91 (-18.26) & 54.35 & 92.12 & 76.19\\
 \#4 &  NL & 87.32 (-6.5) & 52.24 & 99.80 & 51.49 & 72.53 (-18.64) & 89.21 & 99.99 & 50.38 \\
\hline
\end{tabular}
\end{center}
\caption{Analysis of measuring significance of each step of SelNLPL. \#1: SelNLPL. \#2: Deleting SelPL from \#1. \#3: Deleting SelNL from \#1. \#4: Deleting SelNL and SelPL from \#1.}
\label{tab:Ablation study}
\vspace{-3mm}
\end{table*}

\subsection{Results}
\label{sec:Results}
Table~\ref{tab:Generalized} shows the results from Zhang \etal~\cite{zhang2018generalized}, supplemented with our results. An 18-layer ResNet~\cite{he2016deep} was used for FashionMNIST, whereas a 34-layer ResNet~\cite{he2016deep} was used for CIFAR10 and CIFAR100. Our method achieved the best overall accuracy in almost all cases, regardless of CNN architecture, dataset, noise type, or noise ratio. In some cases, our method outperformed the others significantly, up to a maximum of 5\%. Our method only failed to converge when the \textit{symm-exc} noise is 80\%, which can be neglected because such a scenario is unrealistic. It should be noted that Zhang \etal~\cite{zhang2018generalized} referenced the accuracy of the validation data to prevent overfitting to noisy data, whereas our method is advantageous as it did not reference any validation accuracy. The results of CIFAR100 were achieved by the extended version of our method, which is detailed in Section~\ref{sec:Generalization to Number of Classes}.

Table~\ref{tab:Joint} is taken from Tanaka \etal~\cite{tanaka2018joint}. A 32-layer Pre-ResNet~\cite{he2016identity} was used for CIFAR10. Similar to those in Table~\ref{tab:Generalized}, our method outperformed all other comparable methods reported in Tanaka \etal~\cite{tanaka2018joint}, regardless of noise types and ratios. This result is noteworthy because Tanaka \etal~\cite{tanaka2018joint} conducted their experiments by varying a few hyper-parameters depending on both noise type and noise ratio. In realistic cases, this setting is not applicable because the type or ratio of noise is unknown. Our method is excellent as the hyper-parameters do not vary according to noise type and ratio. Furthermore, for fair comparison for the case of \textit{asymm} noise, we matched the parameter settings used for \textit{asymm} noises to those used for 10\% (Joint*) and 30\% (Joint**) \textit{symm-inc} noises, because the amount of noisy data for \textit{asymm} noise cases lies between that of 10\% \textit{symm-inc} and 30\% \textit{symm-inc}. In that case, the overall accuracy of ~\cite{tanaka2018joint} for \textit{asymm} noise changed, making our method superior for all \textit{asymm} noise cases. 

 Table~\ref{tab:Dimensionality} and~\ref{tab:Semi} are taken from ~\cite{ma2018dimensionality} and ~\cite{ding2018semi}, respectively. While Table~\ref{tab:Dimensionality} adopted structure of LeNet5 for MNIST, Table~\ref{tab:Semi} used a 2-layer fully connected network for MNIST and a 14-layer ResNet for CIFAR10. Both tables show our method surpassed most of the other comparable results for all CNN architectures, datasets, noise types and ratios. In some cases, the performance of our method exceeded those of other methods by up to 4$\sim$5\%, demonstrating the superiority of our method. Our method only performed second best for 60\% \textit{asymm} noise in Table~\ref{tab:Semi}, but we believe this is unimportant because such a scenario is unrealistic.
%-------------------------------------------------------------------------
\vspace{0mm}
\section{Analysis}
\label{sec:Analysis}
\vspace{0mm}
\subsection{Generalization to number of classes} % Convergence on CIFAR100
\label{sec:Generalization to Number of Classes}

Our method NL is an indirect learning method using a complementary label. Owing to the nature of NL's optimization process, the amount of convergence depends on class number $c$ in the dataset; as $c$ increases, the training of the CNN becomes slower. Consequently, our method failed to converge on CIFAR100 when training the CNN with the same number of epochs as used when training the CNN with CIFAR10. To overcome and analyze this phenomenon, we observed the gradients resulting from NL (Eq.~\ref{eq:NL}). Here, we consider the gradients associated with clean data points to gain insight into the way we extend the NL method to many class cases. Let us consider a data point with a clean label, implying given $\overline{y}$ always is not true label $y$. By assuming that the CNN is at its initial state, the following probability values in Eq.~\ref{eq:NL} are approximated to be uniform across all classes ($\boldsymbol{p}_i=\frac{1}{c}$). The gradients are approximated as follows:

\begin{equation}
\label{eq:NL gradient}
\begin{split}
& \frac{\partial \mathcal{L}(f,\overline{y})}{\partial f_i} =
\begin{cases}
    \boldsymbol{p}_i \approx \frac{1}{c} & \text{if $i=\overline{y}$} \\
    -\frac{\boldsymbol{p}_{\overline{y}}}{1-\boldsymbol{p}_{\overline{y}}}\boldsymbol{p}_i \approx -\frac{1}{c(c-1)} & \text{if $i \neq \overline{y}$}
\end{cases}
\end{split}
\end{equation}
The detailed outline of Eq.~\ref{eq:NL gradient} is given in the appendix. 

Eq.~\ref{eq:NL gradient} shows that while gradient occurs to reduce the score corresponding to the given $\overline{y}$, a gradient also occurs to enhance the scores corresponding to other remaining classes, including true label $y$. This implies that after training CNN with NL, the gradient received at $y$ is $\frac{1}{c(c-1)}$.

\vspace{-0.5mm}
Suppose we are training the CNN with either 10-class dataset or 100-class dataset. The gradient received at $y$ is $\frac{1}{9*10}$ for 10-class dataset and $\frac{1}{99*100}$ for 100-class dataset. Comparing these two cases, the gradient of the 100-class dataset is 110 times smaller than that of the 10-class dataset.
This analysis implies that, for NL to converge on CIFAR100, it requires more epochs than CIFAR10, approximately up to a factor of 110. However as it requires a significant amount of time to the train CNN, we extend our method to provide multiple random complementary labels for each image. We computed 110 losses on a single data with 110 random $\overline{y}$s (allowing duplication), with only slight increase in time for back-propagation as the 110 losses share features computed for one image.
With this simple extended method, we observed that the CNN could converge on CIFAR100 when trained with the same number of epochs as CIFAR10, and showed general improvement in noisy data classification (Table~\ref{tab:Generalized}). For \textit{symm-exc} noise, we achieved state-of-the-art results. For \textit{asymm} noise, Forward T~\cite{patrini2017making} shows the best performance. However, this is not a fair comparison as it relies on the prior knowledge of the confusion matrix, which summarized the probability of one class being flipped into another under noise. Therefore it can be concluded that our method achieved comparable results excluding Forward T.

In this section, we demonstrated that our method can be generalized to datasets with many class numbers by providing multiple complementary labels for each image.

\subsection{Ablation study}
\label{sec:Ablation Study}
Our paper suggests a novel noisy data classification method, composed of multiple steps: SelNLPL (NL$\rightarrow$SelNL$\rightarrow$SelPL), followed by pseudo labeling for semi-supervised learning. To investigate the strength of each step in SelNLPL, we conducted an analysis that reveals the difference in performance when each step of SelNLPL is omitted from the entire training process. One or more steps are deleted from SelNLPL and then applied to pseudo labeling. Table~\ref{tab:Ablation study} shows all experiments conducted for the analysis, following the experimental settings of Table~\ref{tab:Joint} with 30\% and 50\% \textit{symm-inc} noise. It includes SelNLPL (\#1) and deleting either SelPL (\#2), SelNL (\#3), or both SelNL and SelPL (\#4) from \#1. 

In Table~\ref{tab:Ablation study}, compared to \#1, both \#2 and \#3 show deteriorated performance, while \#4's performance is further decreased (Figure~\ref{fig:SelPLNL_results-50}). 
Although the accuracy drop is quite small on \textit{symm-inc} 30\% noise case, significant degradation in performance is shown in \textit{symm-inc} 50\% noise case, especially for \#3 and \#4, where SelNL is deleted from the training process. Convergence of CNN when training with NL depends heavily on the amount of noise in the training data, being less convergent as the noise ratio increases (Figure~\ref{fig:SelPLNL_results} (b), Figure~\ref{fig:SelPLNL_results-50} (a)). Consequently, SelNL becomes crucial with the increase in noise ratio as it is responsible for ignoring the noisy data having low confidence value. SelNL enhances the overall clean data ratio involved in training, thus yielding better convergence (Figure~\ref{fig:SelPLNL_results} (c), Figure~\ref{fig:SelPLNL_results-50} (b)).

%-------------------------------------------------------------------------
\section{Conclusion}
\label{sec:Conclusion}
We proposed NL for training with noisy data, a learning method of training CNNs that ``input image does not belong to this complementary label.'' This reduces the risk of training CNNs with wrong information as the chances of randomly choosing a complementary label that is not a ground-truth label are high. Furthermore, because PL is faster and more accurate when learning with clean data compared to NL, we developed a new method, SelNLPL, by combining PL and NL, obtaining a superior performance for filtering noisy data from training data. Our study performed successful noisy data classification by using semi-supervised learning (pseudo labeling) based on the filtering result of SelNLPL, achieving state-of-the-art results without tuning our method based on any prior knowledge. 
%-------------------------------------------------------------------------

{\small
\bibliographystyle{ieee_fullname}
\bibliography{main}
}

\end{document}